\documentclass[a4paper,12pt,onecolumn]{article}

\usepackage{FAS}
\usepackage[latin1]{inputenc}
\usepackage[english]{babel} 
\usepackage{graphicx}
\usepackage{hyperref}

\usepackage{amsmath}
\usepackage{amssymb}  
\usepackage{bm}
\usepackage{cite}
\usepackage[tracking]{microtype}
\usepackage{csquotes}
\usepackage{tikz}
\usepackage{wrapfig} 
\usepackage[draft,inline,nomargin,index]{fixme}
\usepackage{mathtools}
\usepackage{comment}
\usepackage[super]{nth}
\usepackage[shortcuts, acronym, toc, nonumberlist=false]{glossaries}
\makenoidxglossaries

\newcommand{\eg}{e.g.,\ }

\usepackage{booktabs}
\usepackage{boldline}
\usepackage{siunitx}
\DeclareSIUnit{\rad}{rad}
   \newacronym
 {pomdp}{POMDP}{Partially-Observable Markov Decision Process}
  \newacronym
 {rl}{RL}{Reinforcement Learning}
  \newacronym
 {il}{IL}{Imitation Learning}
  \newacronym
 {irl}{IRL}{Inverse Reinforcement Learning}
  \newacronym
 {mp}{MP}{Message Passing}
  \newacronym
 {av}{AV}{automated vehicle}
   \newacronym
 {ppo}{PPO}{Proximal Policy Optimization}
   \newacronym
 {bc}{BC}{Behavior Cloning}
   \newacronym
 {mlp}{MLP}{multi-layer perceptron}
   \newacronym
 {gae}{GAE}{Generalized Advantage Estimation}
    \newacronym
 {gan}{GAN}{Generative Adversarial Network}
    \newacronym
 {airl}{AIRL}{Adversarial Inverse Reinforcement Learning}
    \newacronym
 {rmse}{RMSE}{Root Mean Squared Error}
    \newacronym
 {sl}{SL}{Supervised Learning}

\begin{document}
%
\title{Conditional Prediction by Simulation for Automated Driving}

\renewcommand\thefootnote{}  
\footnotetext{This work is a result of the joint research project STADT:up (19A22006E). The project is supported by the German Federal Ministry for Economic Affairs and Climate Action (BMWK), based on a decision of the German Bundestag. The author is solely responsible for the content of this publication.}
\renewcommand\thefootnote{\arabic{footnote}}  

\author{
    Fabian Konstantinidis\thanks{Pre-Development of Automated Driving, CARIAD SE, 38440 Wolfsburg, Germany. (e-mail: firstname.lastname@cariad.technology)} \textsuperscript{,}%
    \thanks{Institute of Measurement and Control Systems, Karlsruhe Institute of Technology (KIT), 76131 Karlsruhe, Germany. (e-mail: fabian.konstantinidis@partner.kit.edu, stiller@kit.edu)} , 
    Moritz Sackmann\footnotemark[1] , 
    Ulrich Hofmann\footnotemark[1] , \\
    Christoph Stiller\footnotemark[2]
}

%
\date{}

\maketitle \thispagestyle{empty}

\begin{abstract}
Modular automated driving systems commonly handle prediction and planning as sequential, separate tasks, thereby prohibiting cooperative maneuvers. To enable cooperative planning, this work introduces a prediction model that models the conditional dependencies between trajectories. For this, predictions are generated by a microscopic traffic simulation, with the individual traffic participants being controlled by a realistic behavior model trained via Adversarial Inverse Reinforcement Learning. By assuming various candidate trajectories for the automated vehicle, we generate predictions conditioned on each of them. Furthermore, our approach allows the candidate trajectories to adapt dynamically during the prediction rollout. Several example scenarios are available at \href{https://conditionalpredictionbysimulation.github.io/}{https://conditionalpredictionbysimulation.github.io/}.
\end{abstract}


\section{Introduction}
Predicting the future trajectories of surrounding traffic participants plays an essential role in automated driving. By anticipating future movements of nearby agents, such as vehicles and vulnerable road users, an \gls{av} can better plan maneuvers, reduce the risk of collisions, and ensure smoother interactions with other road users.

Although existing approaches, \eg \cite{Seff23MotionLM, Shi24MTR++, Wagner2024SceneMotion}, effectively predict the future movements of individual traffic participants, they limit an \gls{av} to a reactive planning strategy, assuming that the predictions of surrounding vehicles remain unaffected by the \gls{av}'s planned actions. In highly interactive situations, this often leads to the \emph{freezing robot problem} \cite{Trautmann10UnfreezingRobot}, where the \gls{av}, unable to engage in cooperative planning, simply stops to avoid potential collisions. For example, when it is unable to merge in dense traffic because the predictions of surrounding vehicles do not react to the \gls{av}'s plan.

One approach to resolving this is to condition the prediction on the \gls{av}'s plan, often referred to as \emph{conditional inference} \cite{Tolstaya21ConditionalInference}. This enables holistic planning: Repeatedly selecting candidate trajectories and predicting their impact on surrounding vehicles until a sufficiently good trajectory is found. A straightforward approach to this involves extending the prediction model to incorporate the \gls{av}'s planned trajectory, improving the prediction by considering the \gls{av}'s influence. However, this requires the \gls{av}'s trajectory to be fixed during prediction, preventing truly interactive planning. We argue that an effective prediction model should capture the bidirectional interactions by conditioning the prediction on the plan of the \gls{av} while simultaneously allowing the plan to adapt in response to the evolving predictions, \eg when planning stepwise in a tree structure \cite{Chen23TreeBasedPlanning}.

\begin{figure}[t]
\centering
\includegraphics[width=12.4cm]{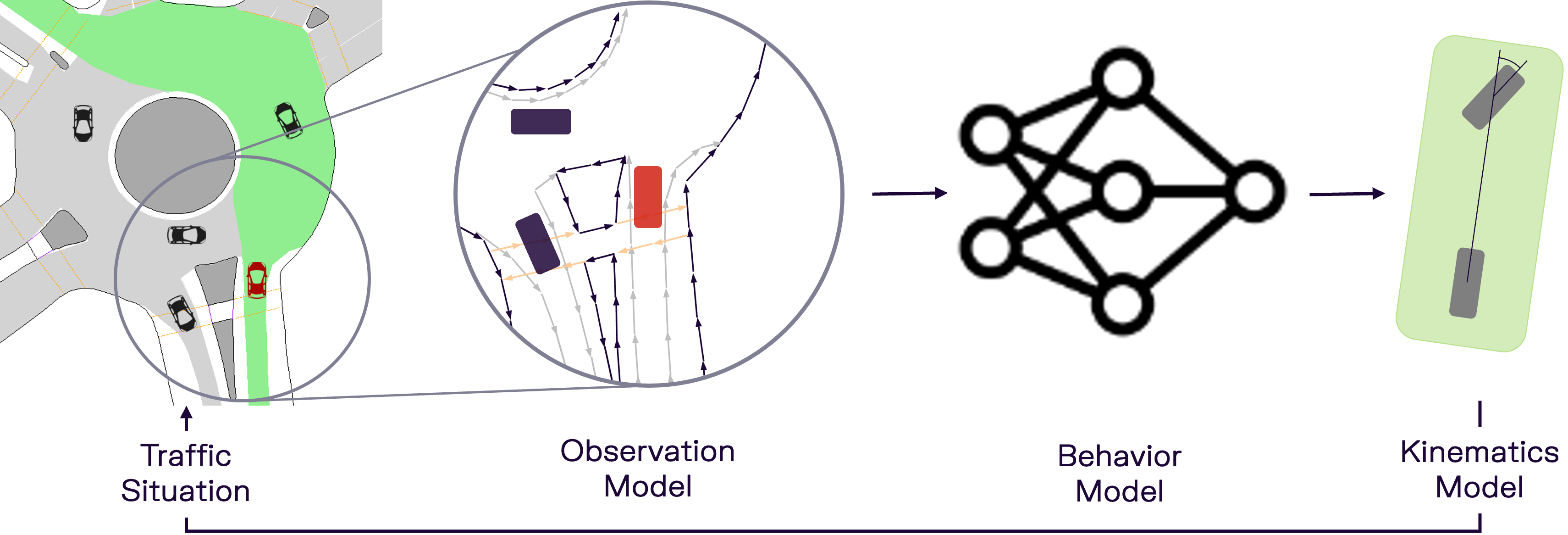}
\vspace{-12pt}
\caption{Single simulation step: Each vehicle observes the traffic situation locally, selects an action based on the observation, and executes it using the kinematics model.} 
\label{fig:sim}
\vspace{-12pt}
\end{figure}

In this work, we realize such a system by learning a reactive behavior policy and utilizing it for simulating the surrounding vehicles in a closed-loop simulation. 
As depicted in Figure \ref{fig:sim}, our framework predicts the evolution of traffic situations by stepwise simulating the movements of all surrounding vehicles using the learned behavior policy until the prediction horizon is reached. 
At each simulation step, the model processes an observation describing the traffic situation from each agent's perspective and executes appropriate control actions accordingly. This process is conducted stepwise and independently for each target vehicle, allowing agents to respond to each other's movements in the subsequent simulation steps. This approach not only facilitates scene-consistent predictions but also allows the \gls{av} to adapt its planned trajectory dynamically during the prediction rollouts.

Realistically simulating driving behavior requires a behavior model, typically derived from data-driven methods like \gls{rl} \cite{Konstantinidis23} or \gls{il} \cite{Sackmann22AIRL}. While \gls{rl} enhances robustness, it requires the definition of a reward signal that describes realistic behavior, which is a complex task. In contrast, \gls{bc}, an \gls{il} method, learns from expert demonstrations using supervised learning but suffers from \emph{covariate shift} \cite{Spencer21}, where compounding errors during inference lead to situations differing from the training data, resulting in undefined behavior.

Similar to our previous work \cite{Konstantinidis24AIRL}, we address these limitations by employing \gls{airl} \cite{Fu17AIRL} to learn the behavior model. \gls{airl} combines \gls{rl} and \gls{il} by reconstructing the reward function that best explains the behavior of real- world drivers, and simultaneously learning a behavior model that maximizes this reward.

\textbf{Contributions}: 
The core contribution of this work is the integration of flexible behavior models, learned via \gls{airl}, into a simulation framework to enable conditional motion prediction. Specifically, this study emphasizes the prediction of surrounding vehicles conditioned on a predefined plan for the \gls{av}, thereby capturing the bidirectional interactions between the \gls{av} and its surrounding agents. As we focus on prediction, we maintain fixed paths for the \gls{av} during prediction; however, the proposed stepwise approach enables the planning algorithm to adapt its strategy during the prediction rollout. In addition, we apply the model to various traffic situations with distinct road layouts. 

\section{Method}
Traffic situations are simulated by executing a learned behavior model for every agent in the scene simultaneously. 
From the perspective of an individual agent, surrounding drivers are treated as part of the environment. When driving, the uncertainty arising from their unknown driving characteristics (\eg individual driving styles or intentions) adds complexity to the sequential decision-making process, often formulated as a \gls{pomdp}. Formally, a \gls{pomdp} is characterized by the tuple $(S, O, A, T, R, \Omega, \gamma)$. In this framework, the agent cannot directly observe the true state $s \in S$ but instead is limited to a (noisy) observation $o \in O$ of the environment, determined by the observation model $\Omega: S \rightarrow O$ mapping from states to observations. Upon executing an action $a \in A$, the state of the environment is updated stochastically according to the transition probability density $T: S \times A \times S \rightarrow \left[ 0, \infty \right[$. Additionally, the agent receives a numerical reward defined by the reward function $R: S \times A \rightarrow \mathbb{R}$ as well as an observation of the new state of the environment. The discount factor $\gamma \in \left[0, 1 \right[$ balances the trade-off between immediate and future rewards. The solution of a \gls{pomdp} is the optimal policy $\pi^*: O \times A \rightarrow \left[0, \infty \right[$ mapping from observations to distributions over actions that maximize the expected cumulative reward over time $J(\pi) = \mathop{\mathbb{E}}_{\pi} \left[ \sum_{k=0}^\infty \gamma^k r_k \right]$, where $r_k = R(s_k, a_k)$.

\subsection{Reinforcement Learning}
A common approach to maximizing $J(\pi)$ is through \gls{rl}, where the agent interacts with a (simulated) environment via trial and error to discover effective actions. 
This involves two alternating steps: 1) The agent collects a new set of experiences $E = \{ e_1, \dots, e_M \}$, where $e_k = \left(o_k, a_k, r_k \right)$ represents one experience obtained by executing a single step in the environment. 2) The experiences gathered are used to adjust the policy, promoting actions that lead to higher rewards and vice versa for actions that lead to lower rewards. These steps are repeated until a sufficiently good policy is found. 

Commonly, a parameterized policy $\pi_\theta$ is used, where $\theta$ denotes the trainable parameters of the policy. During the policy update, these parameters are updated according to
\begin{equation}
    \theta \leftarrow \theta + \alpha \frac{1}{M} \sum_{e_k \in E} A(o_k, a_k) \triangledown_\theta \log \pi_\theta (a_k \mid o_k),
\end{equation}
where $\alpha$ is the learning rate. Here, the gradients of the policy $\triangledown_\theta \log \pi_\theta(a_k \mid o_k)$ are weighted by the advantage $A(o_k, a_k)$ denoting how much the action $a_k$ is better ($A(o_k, a_k) > 0$) or worse ($A(o_k, a_k) < 0$) when making an observation $o_k$ compared to acting according to $\pi_\theta$.

The difficulty in our setting is that we are learning a policy for every vehicle in the scene simultaneously, making the environment non-stationary from an agent's point of view. To mitigate this issue, similar to our previous work \cite{Konstantinidis23}, every agent uses a copy of the same shared policy, allowing the use of single-agent methods to solve this multi-agent task. 
For our experiments, we use \gls{gae} \cite{Schulmann15} for obtaining an estimate of the advantage $\hat{A}(o_k, a_k)$ and the \gls{ppo} \cite{Schulmann17} algorithm for learning the driver model.

\subsection{Adversarial Inverse Reinforcement Learning}
Our goal is to predict human driving behavior by learning and executing a policy that accurately models human driving behavior.
Utilizing \gls{rl} for learning the policy requires the definition of a reward function that accurately captures the incentive structure of real-world drivers. For human driving, finding such a reward function is a time-consuming and tedious task. However, it is easy to demonstrate the desired behavior in the form of a set of demonstrations recorded in real traffic. 

One way to automate the process of defining a reward function is \gls{airl}, where a surrogate reward signal is reconstructed, which explains the demonstrated behavior. Utilizing the reconstructed reward for learning the policy model yields policies that mimic expert demonstrations, given as $\mathcal{D}=\{ (o_1, a_1), (o_2, a_2), \dots \}$. 
Specifically, \gls{airl} combines \gls{rl} with the ideas of \gls{gan} and applies them to the task of \gls{il}. The policy $\pi_\theta$ is still learned via \gls{rl} maximizing a reward signal, but the reward signal is now approximated by a discriminator model. The discriminator model $D_\phi$ tries to distinguish generated samples from demonstrated ones by assigning higher scores to more realistic samples, whereas the generator model (the trainable policy $\pi_\theta$) tries to fool the discriminator by generating samples matching the distribution of the demonstrated data. These adversarial objectives can be modeled with the following two-player minimax game:
\begin{equation}\label{eq:minimax}
    \min_\phi \max_\theta \biggl[ -\mathop{\mathbb{E}}_{(o, a) \sim \mathcal{D}} \left[ \log \left( D_\phi (o, a) \right) \right]  -\mathop{\mathbb{E}}_{(o, a) \sim \pi_\theta} \left[ \log \left( 1 - D_\phi (o, a) \right) \right] \biggr],
\end{equation}
where the discriminator assigns the probability $D_\phi (o, a) \in \left[0, 1 \right]$ to the observation-action pair being real. By imposing a special structure on the discriminator
\begin{equation}\label{eq:airl_probab_exp}
    D_\phi(o, a) = \frac{\exp \left(f_\phi (o, a) \right)}{\exp \left( f_\phi (o, a) \right) + \pi \left(a \mid o \right)},
\end{equation}
a surrogate reward signal is reconstructed, where samples with high rewards are exponentially more likely than samples with low rewards. This structure corresponds to the odds ratio between the policy $\pi \left(a \mid o \right)$ and the exponentiated reward distribution $\exp \left( f_\phi (o, a) \right)$. The discriminator is trained by minimizing the binary cross-entropy loss in (\ref{eq:minimax}) and the generator via \gls{rl} with the surrogate reward function 
\begin{equation}\label{eq:airl_surrogate}
    \begin{aligned}
        \tilde{r}(o, a) &= \log \left( D_\phi (o, a) \right) - \log \left( 1 -  D_\phi (o, a) \right) + c \\
        &= f_\phi(o, a) - \log \pi \left(a \mid o \right) + c,
    \end{aligned}
\end{equation}
where $c$ is a constant reward and $- \log \pi \left(a \mid o \right)$ rewards policies for higher entropy, thus promoting exploration during training and robustness to action noise.

\textbf{\emph{Modifications}}:
In general, \gls{airl} is a domain-agnostic method. However, when applied to learning a driver model, as proposed in \cite{Sackmann22AIRL, Konstantinidis24AIRL}, two modifications are required: 1) In the original implementation \cite{Fu17AIRL}, $c=0$ is used in (\ref{eq:airl_surrogate}), leading to a negative expected value for the surrogate reward $\tilde{r}(o, a)$, as the discriminator is typically able to classify correctly. In our setting, this resulted in suicidal vehicles leaving the track immediately to avoid the pain of continuously receiving negative rewards. To alleviate this issue, we set $c=5$, thus promoting survival without changing the optimal behavior with respect to the discriminator model. 
2) While during \gls{rl} training, a large variance in the actions drawn from the policy is crucial to gather a diverse set of experiences, it allows the discriminator to easily detect generated samples. Therefore, to smoothen the decision boundary, during discriminator training, we add random noise to the actions executed by the experts matching the standard deviation of the learned policy.

\subsection{Model Details}
To ensure accurate predictions, the model must understand both its environment and the interactions between traffic participants. To this end, we employ a flexible graph-based observation and model architecture, similar to \cite{Konstantinidis24AIRL}, which is particularly effective for handling complex road topologies. 
Specifically, we use agent-centric observations that capture both agents and road elements within a $\SI{30}{\meter}$ radius.  
Agents are described by their size (width and length), position, heading, velocity, and current speed limit: $\mathbf{x} = \left[ \mathbf{x}_\mathrm{size}, \mathbf{x}_\mathrm{pos}, \mathbf{x}_\mathrm{head}, x_\mathrm{vel}, x_\mathrm{limit} \right]^\intercal$. Road elements, such as road markings and boundaries, are represented as sets of vectors, with each vector being defined by its start and end points, a one-hot type encoding, and a binary flag indicating whether it is part of the assigned route: $\mathbf{v} = \left[ \mathbf{v}_\mathrm{start}, \mathbf{v}_\mathrm{end}, \mathbf{v}_\mathrm{type}, v_\mathrm{route} \right]^\intercal$. Vectors corresponding to the same road element are grouped into polylines, which are then interconnected with the agent nodes, forming a higher-level interaction graph. 

We use a similar model for both the policy and the discriminator, with the primary distinction being in the decoder. In our model, polylines of varying length are encoded through multiple layers of \gls{mp}, as expressed by:  
\begin{equation}
    \mathbf{v} \leftarrow f_\mathrm{rel} \left(g_\mathrm{enc} \left(  \mathbf{v} \right), f_\mathrm{agg} \left( \{g_\mathrm{enc} \left( \mathbf{v}_l \right)\}_{l=1}^L \right)  \right),
\end{equation} 
with $g_\mathrm{enc}(\cdot)$ being an \gls{mlp}, $f_\mathrm{agg}(\cdot)$ an element-wise max-pooling operation and $f_\mathrm{rel}$ a simple concatenation. Here, $v_l \in \{ v_1, v_2, \cdots, v_L \}$ denotes one of the $L$ vectors forming the polyline. Lastly, the polyline embeddings $\mathbf{q}$ are obtained by applying an element-wise max-pooling operation over all polyline vectors. 
Agent feature vectors are encoded into the same embedding space using a simple \gls{mlp} returning the agent embedding $\mathbf{z} = \mathrm{MLP}(\mathbf{x})$. To make the target agent aware of its surroundings, we use a cross-attention operation $\mathbf{z}_\mathrm{target} \leftarrow \mathrm{CA}\left(\mathrm{Q}\!: \mathbf{z}_\mathrm{target}, \; \mathrm{KV}\!: \left( \mathbf{z}, \mathbf{q}\right) \right)$ \cite{Vaswani17} between the target agent embedding $\mathbf{z}_\mathrm{target}$ and the combined embeddings $\left( \mathbf{z}, \mathbf{q}\right)$. Finally, the decoder, implemented as another \gls{mlp}, maps the interaction- and map-aware embedding $\mathbf{z}_\mathrm{target}$ to the mean and standard deviation of the next acceleration and steering angle for the generator, and the expert probability $D_\phi (o, a)$ for the discriminator. As the discriminator classifies observation-action pairs, the executed action is concatenated to its decoder input.

\subsection{Conditional Prediction by Simulation}\label{sec:conditional_prediction_by_simulation}

The primary motivation of this work is not to demonstrate how \gls{airl} can be used to learn behavior models that imitate human drivers, as this has already been shown in existing works, \eg \cite{Sackmann22AIRL, Konstantinidis24AIRL}. Instead, we focus on demonstrating how a behavior model learned via \gls{airl} can be used to predict the future evolution of traffic situations conditioned on the planned movement of an \gls{av}. This is achieved by embedding the behavior model within a closed-loop simulation framework. In our simulation framework, as illustrated in Figure \ref{fig:sim}, each vehicle is assigned a predefined route. When predicting trajectories online, this information must be inferred beforehand by estimating the likelihood of possible route hypotheses, as done in \cite{Schulz18}. 

\begin{wrapfigure}{r}{0.48\textwidth} 
\centering
\includegraphics[width=\linewidth]{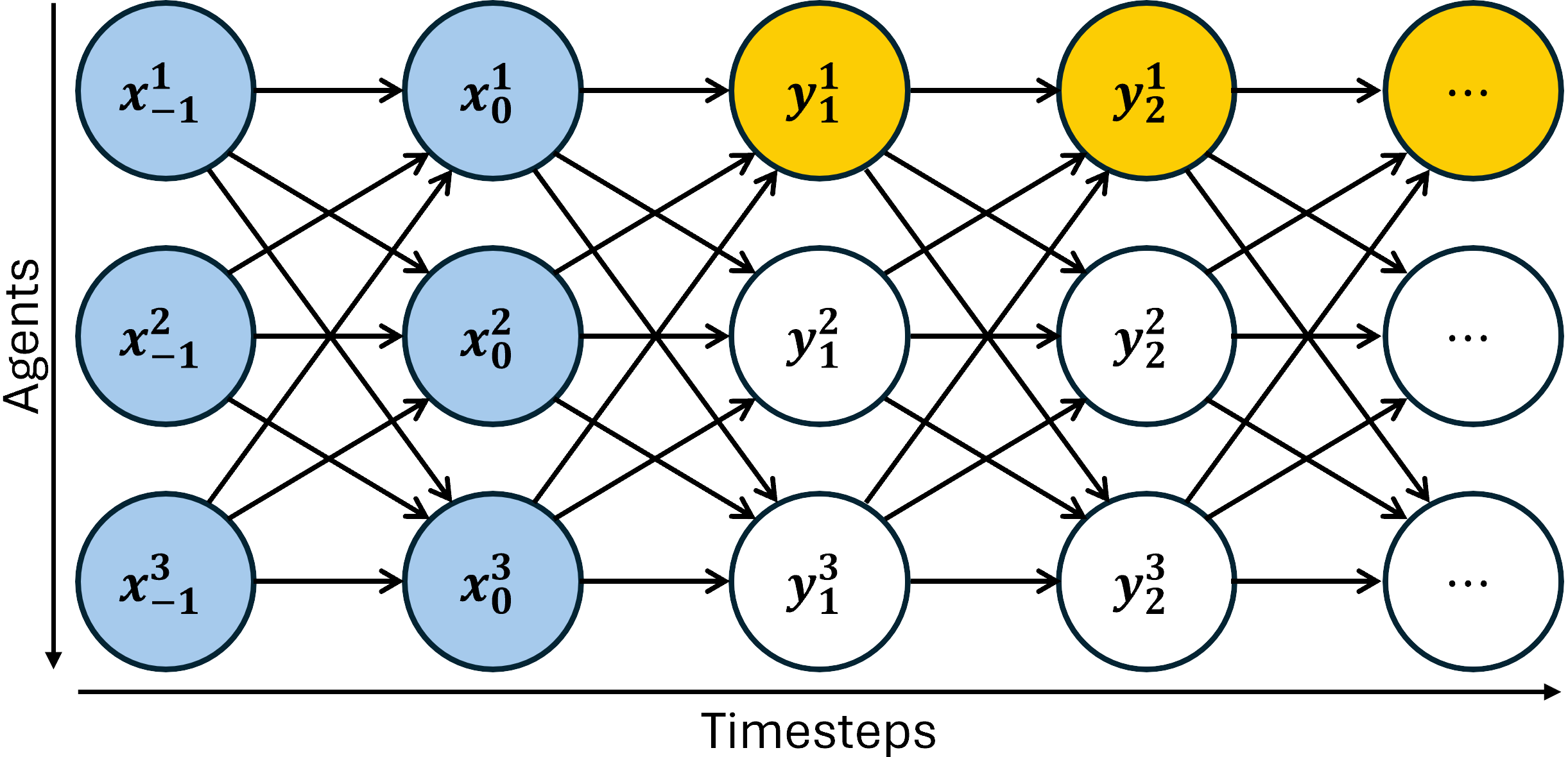}
\vspace{-25pt} 
\caption{Conditional prediction rollout with past states (blue), \gls{av}'s plan (orange), and predicted states (white).} 
\label{fig:prediction_rollout}
\vspace{-8pt} 
\end{wrapfigure}

As illustrated in Figure \ref{fig:prediction_rollout}, the predictions unfold step-wise, with each transition corresponding to a single simulation step (see Figure \ref{fig:sim}). When conditionally predicting the evolution of a traffic situation, the \gls{av} first generates a planned trajectory aligned with its driving objective (represented as the orange sequence of states). At each simulation step, the \gls{av}'s position is updated according to this plan, while the movements of surrounding vehicles are predicted using the learned behavior model. Note that each predicted state depends on the previous state of all vehicles, including itself and the \gls{av}. This allows the vehicles to react to each other's predicted movement in subsequent steps, thereby fostering interaction in the prediction. Additionally, by mutually conditioning prediction and planning this way, the \gls{av} can not only assess how surrounding vehicles influence its plan but simultaneously how its planned trajectory affects their predictions. 

In general, the planned movement of the \gls{av} is not required to be fixed during the prediction rollout. Instead, it can adapt dynamically in response to the predicted reactions of surrounding agents. Such a reactive planner could be realized by using a behavior model similar to the one used for prediction. However, prediction and planning have distinct requirements. While the prediction model must accurately capture human driving behaviors (\eg tailgating or reckless driving), the \gls{av} must prioritize safe and conservative driving. Defining a safe and cooperative behavior planner is beyond the scope of this work. Therefore, we leave this for future research and use manually defined fixed paths for the \gls{av} during prediction.

\section{Experiments}

\begin{wraptable}{tr}{0.5\textwidth} 
  \centering
  \vspace{-12pt} 
  \resizebox{0.5\textwidth}{!}{ \begin{tabular}{l l}
    \hline
    \textbf{Parameter} & \textbf{Value} \\
    \hline
    Agent feature encoder & $\left(8, 64, 64 \right)$ \\
    Number of \gls{mp} layers & $3$ \\
    \gls{mp} \gls{mlp} (layer 1) & $\left(11, 64, 32 \right)$ \\
    \gls{mp} \gls{mlp} (layer 2 \& 3) & $\left(64, 64, 32 \right)$ \\
    Interaction module & cross-attention \\
    Policy decoder & $\left(64, 64, 4\right)$ \\  
    Discriminator decoder & $\left(64\! +\! 2, 64, 1\right)$ \\
    Activation function & ReLU \\
    Batch Size & $1024$ \\
    Optimizer & Adam\cite{Kingma14} \\
    Policy learning rate & $2\cdot 10^{-4}$ \\
    Discriminator learning rate & $1\cdot 10^{-4}$ \\
    Discount factor $\gamma$ & $0.95$ \\
    \gls{gae} $\lambda$ \cite{Schulmann15} & $0.95$ \\
    \gls{ppo} clip range & $0.2$ \\
    \hline
  \end{tabular}}
  \vspace{-10pt} 
  \caption{Training Parameters}
  \label{tab:training_parameter}
  \vspace{-12pt} 
\end{wraptable}

\textbf{\emph{Dataset}}: 
For our experiments, we use the publicly available INTERACTION dataset \cite{Zhan19}, which contains recordings from $11$ locations, including roundabouts, unsignalized intersections, and merging scenarios. The data consists of \num{36279} vehicles, recorded at a frequency of \SI{10}{Hz} over a total duration of \SI{931}{min}. The amount of available data varies significantly across the locations due to differences in traffic density and recording times. Hence, to address this imbalance, we downsampled the data to balance sample sizes across locations during training. For validation and final testing, we use \SI{20}{\%} and \SI{30}{\%} of the recordings per location, respectively. The data is divided into 10-second situations, excluding those where vehicles could not be assigned a route (\eg due to illegal turns). Modeling vulnerable road users is beyond the scope of this work; thus, the corresponding tracks are omitted.

\textbf{\emph{Training}}: 
Table \ref{tab:training_parameter} outlines the network shapes and training parameters. For the baseline, a model is trained using \gls{bc} by minimizing the negative log-likelihood of the expert actions under the predicted action distribution. The \gls{airl} models are trained for \num{10000} epochs, requiring \SI{43}{h} on a single RTX8000 48GB GPU. During each epoch, training situations are randomly initialized, featuring an average of $880$ simulated vehicles. The traffic situations used for validation and final testing include \num{1019} and \num{2080} vehicles, respectively, evenly distributed across the available locations. Each situation is simulated for $50$ timesteps with $\Delta t = 0.2\mathrm{s}$ between consecutive timesteps. Vehicles that reach the end of their assigned route, leave the track, or collide are removed from the scene and the simulation is continued with the remaining vehicles. For final testing, we use the model with the lowest \gls{rmse} after \SI{10}{s} on the validation data.

\subsection{Prediction Performance}

\begin{wrapfigure}{r}{0.5\textwidth} 
\centering
\vspace{-12pt}
    \includegraphics[width=\linewidth]{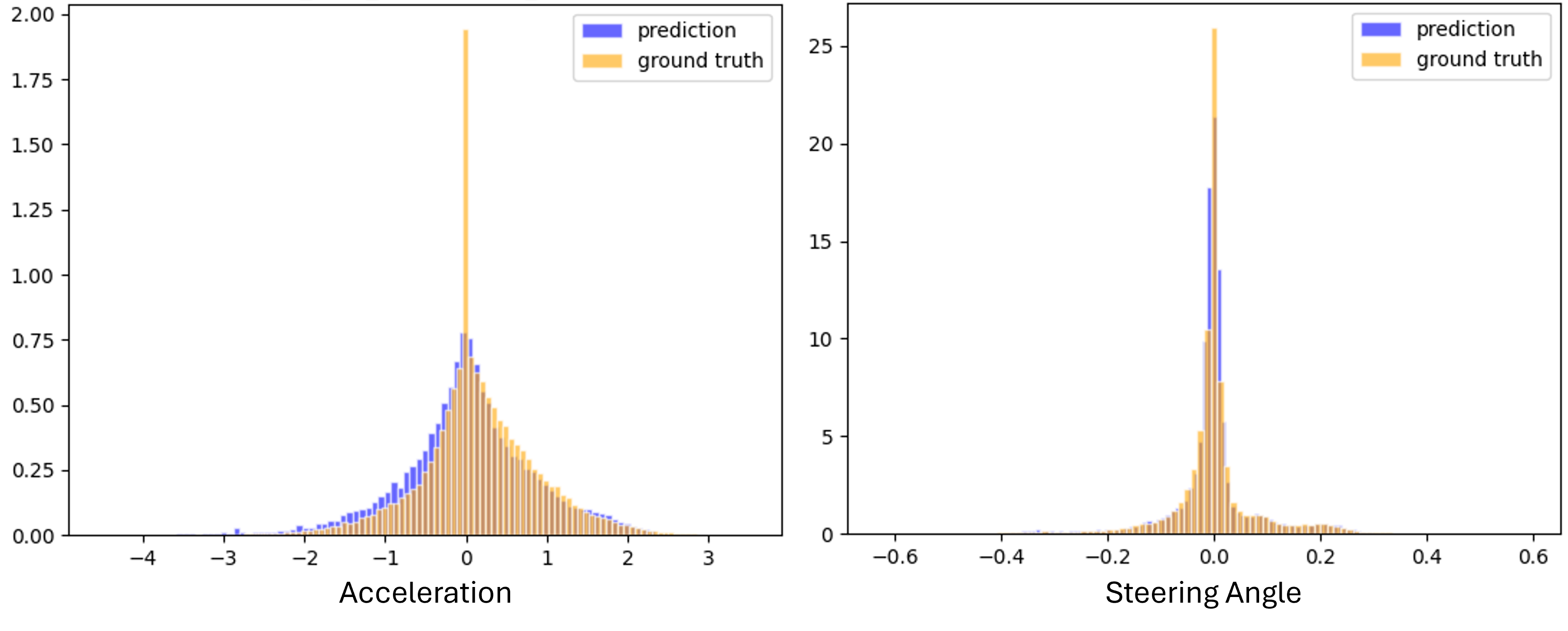}
    \vspace{-30pt}
    \caption{Normalized histograms of executed actions.}
    \label{fig:histo}
    \vspace{-12pt} 
\end{wrapfigure}

For the final evaluation, we simulate each test situation for a duration of \SI{10}{s} by employing the learned behavior policies for all agents in the scene. The initial situations are taken from the ground truth test data. Subsequently, the chosen actions as well as the generated trajectories are compared against those of the corresponding ground truth vehicles.

\begin{wraptable}{r}{0.5\textwidth} 
    \centering
    \vspace{0pt}
    \resizebox{0.5\textwidth}{!}{ 
        \begin{tabular}{l c c c}
            \hline
            \textbf{Model} & \textbf{RMSE} & \textbf{Collision} & \textbf{Off-Track} \\
            \hline
            \gls{bc} & $14.35_{\pm0.42}\text{m}$ & $10.61_{\pm1.87}\%$ & $5.88_{\pm0.85}\%$ \\
            \gls{airl} & $13.63_{\pm0.34}\text{m}$ & $0.65_{\pm0.26}\%$ & $0.23_{\pm0.07}\%$ \\
            \hline
        \end{tabular}
    }
    \vspace{-6pt}
    \caption{Prediction performance after \SI{10}{s}.}
    \label{tab:pred_performance}
    \vspace{-10pt} 
\end{wraptable}

First, we assess the model's ability to reliably imitate human driving behavior by comparing its chosen actions and generated trajectories with those of the corresponding ground truth vehicles. 
Figure \ref{fig:histo} shows the chosen actions of the learned behavior model (blue) and the corresponding ground truth vehicles (orange) in the test situations. It can be seen that the selected actions largely overlap, indicating that the policy effectively captures the correct distribution of actions. The main discrepancy can be seen in the accelerations: the predicted vehicles exhibit slightly more negative accelerations, whereas the ground truth vehicles show a peak at \SI{0}{m/s^2}. This can be explained by the preprocessing of the dataset, where for vehicles at a standstill (\eg waiting at a yield line) both their acceleration and steering angle were set to zero.

Table \ref{tab:pred_performance} presents the prediction performance of the models trained using \gls{bc} and \gls{airl}. Each cell represents the mean and standard deviation of seven models trained with different random seeds. After predicting for \SI{10}{s}, both models demonstrate a strong performance in terms of \gls{rmse}, with the \gls{airl} model outperforming the \gls{bc} model. 
However, when looking at the collision and off-track rates, the \gls{airl} models show significantly greater robustness. This improvement can be attributed to the models being trained in simulation, where the model can explore and learn from actions that are not present in the dataset. The result is a realistic behavior model that, when executed in a closed-loop simulation, generates accurate and scene-consistent predictions. 

\textbf{Conditional Prediction}: 
Furthermore, we want to demonstrate how the learned behavior model can be used for making conditional predictions. As outlined in Section \ref{sec:conditional_prediction_by_simulation}, we modify the planned trajectories for individual vehicles and predict the remaining vehicles. Similarly, an \gls{av} could query the prediction model with different planned trajectories for itself and select a plan based on the corresponding predictions. Multiple example scenarios are available at \href{https://conditionalpredictionbysimulation.github.io/}{https://conditionalpredictionbysimulation.github.io/}.

\begin{figure}[b!]
\vspace{-9pt}
\centering
\includegraphics[width=0.95\textwidth]{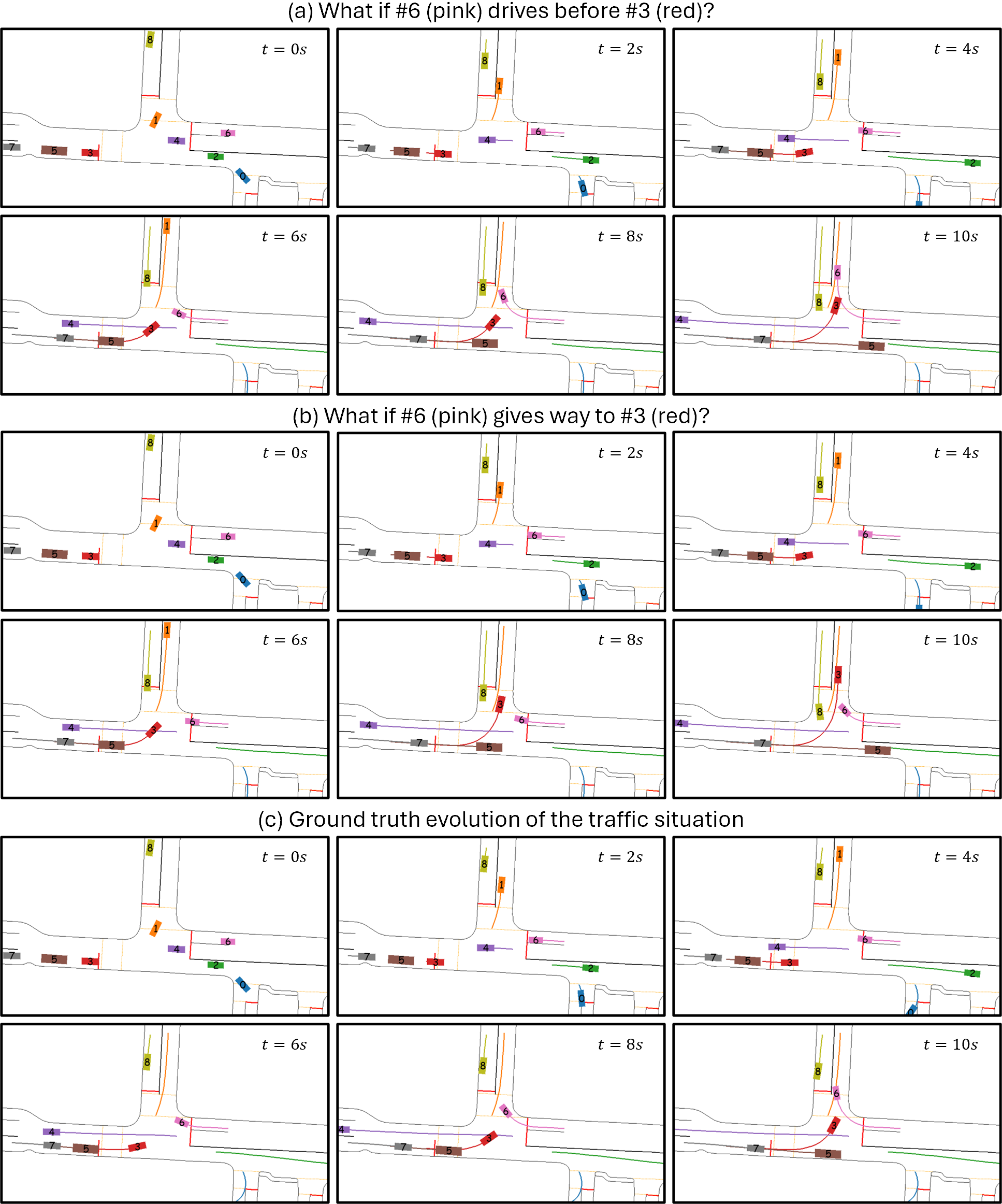}
\vspace{-12pt}
\caption{Demonstration of a conditional prediction.} 
\vspace{-8pt}
\label{fig:conditional_prediction_example}
\vspace{-24pt}
\end{figure}

An example situation is shown in Figure \ref{fig:conditional_prediction_example}. The intersection shown is an all-way stop intersection, where each vehicle is required to stop before entering the intersection. Priority is determined by the order of arrival, resulting in significant interactions between traffic participants. 
In this scenario, the pink vehicle (ID 6) intends to turn right, while the oncoming red vehicle (ID 3) intends to turn left towards the same exit. Assuming that ID 6 is an \gls{av}, it must decide whether to proceed before ID 3 or to yield to it. To evaluate the consequences of both choices, we predict the traffic situation for each of them separately. 
The original prediction of the model is shown in Figure \ref{fig:conditional_prediction_example}(a). This prediction is derived by applying the learned behavior model to all vehicles in the scene. To simulate the conditioning on the alternative future trajectory of ID 6, its executed accelerations are simply replaced by constant breaking for the first \SI{5}{s}. The resulting conditional prediction is shown in Figure \ref{fig:conditional_prediction_example}(b). For the first \SI{2}{s}, both predictions evolve similarly. However, then the predictions start to diverge: Due to ID 6 entering the intersection in \ref{fig:conditional_prediction_example}(a), ID 3 must decelerate and give way to ID 6. Conversely, in \ref{fig:conditional_prediction_example}(b), ID 6 stops, allowing ID 3 to cross the intersection unhindered. As a result, ID 3 advances farther during the prediction, allowing ID 8 to continue driving earlier. The other vehicles' predictions remain largely unaffected, as the model has learned that they are not influenced by the changed trajectory of ID 6.

The corresponding ground truth evolution of the situation is shown in Figure \ref{fig:conditional_prediction_example}(c). There, the vehicle with ID 6 indeed turns first, and the situation unfolds similarly to prediction \ref{fig:conditional_prediction_example}(a). 
Notably, the policy exhibits driving dynamics closely resembling real-world drivers, including comparable velocities, accelerations, and also nuanced driving behavior like cutting corners. 
The main difference in the prediction lies in the duration vehicles stop at a stop line (depicted in red). For instance, the brown vehicle (ID 5) stops for a shorter time in the prediction than its ground truth counterpart.  
This can be explained by the latency between deciding and continuing to drive, as well as the subjective nature of stopping behavior: some drivers stop for several seconds, while others pause only briefly.

\section{Conclusion}
This work presents an autoregressive prediction model realized as a simulation framework that executes learned behavior models for all predicted vehicles. The proposed behavior model uses a flexible graph representation as input and is trained using the \gls{airl} method. 
We evaluated our approach on the INTERACTION dataset, which comprises a variety of traffic scenarios with diverse traffic densities and road layouts. The results show that the model generates realistic driving behaviors, resulting in robust and scene-consistent predictions. Furthermore, we assumed individual vehicles to be \gls{av}s and explored the impact of different candidate trajectories for them on the prediction outcome, highlighting the model's ability to evaluate different planning strategies. 

While this study focused on prediction, with manually generated plans for the \gls{av}, future work could integrate a more advanced planning algorithm to fully leverage the proposed prediction framework.

\begingroup
\fontsize{11.47}{11.4}\selectfont

\section*{Acknowledgment}
The authors would like to thank Matthias Dingwerth and Matthias Steiner for their support and contributions to the development of the simulation framework.

\vspace{-4pt}

\bibliographystyle{IEEEtran} 
\bibliography{library.bib}
\endgroup

\end{document}